\documentclass[letterpaper, 10 pt, conference]{IEEEformats/ieeeconf}  

\IEEEoverridecommandlockouts                              

\usepackage{graphics} 
\usepackage{epsfig} 
\usepackage{mathptmx} 
\usepackage{times} 
\usepackage{amsmath} 
\usepackage{amssymb}  
\usepackage{mathtools}
\usepackage{subcaption}
\usepackage{float}
\usepackage{caption}
\usepackage{csquotes}
\usepackage{siunitx}
\title{\LARGE \bf
	Deep RADAR Inverse Sensor Models for Dynamic Occupancy Grid Maps*}
\author{Zihang Wei$^{1}$, Rujiao Yan$^{2}$, Matthias Schreier$^{2}$
	\thanks{*Z. Wei and R. Yan contributed equally.}
	\thanks{$^{1}$ Z. Wei finished this work during his internship at Continental Autonomous Mobility Germany GmbH, Frankfurt a. M., Germany. 
		{\tt\small zihang.wei@rwth-aachen.de}}%
	\thanks{$^{2}$R. Yan and M. Schreier are with Continental Autonomous Mobility Germany GmbH, Frankfurt a. M., Germany.   
		{\tt\small \{rujiao.yan, matthias.schreier\}@continental.com}
	}%
}
\begin{document}
	\maketitle
	\thispagestyle{empty}
	\pagestyle{empty}
	\begin{abstract}
		To implement autonomous driving, one essential step is to model the vehicle environment based on the sensor inputs. RADARs, with their well-known advantages, became a popular option to infer the occupancy state of grid cells surrounding the vehicle. To tackle data sparsity and noise of RADAR detections, we propose a deep learning-based Inverse Sensor Model (ISM) to learn the mapping from sparse RADAR detections to polar measurement grids. Improved LiDAR-based measurement grids are used as reference. The learned RADAR measurement grids, combined with RADAR Doppler velocity measurements, are further used to generate a Dynamic Grid Map (DGM). Experiments in real-world high-speed driving scenarios show that our approach outperforms the hand-crafted geometric ISMs. In comparison to state-of-the-art deep learning methods, our approach is the first one to learn a single-frame measurement grid in the polar scheme from RADARs with a limited Field of View (FOV). The learning framework makes the learned ISM independent of the RADAR mounting. This enables us to flexibly use one or more RADAR sensors without network retraining and without requirements on 360° sensor coverage.	
	\end{abstract}
	\section{INTRODUCTION}
	Environmental perception, modelling and fusion are key prerequisites to make safe decisions in high-level tasks of autonomous driving, for recent surveys, see \cite{Schreier+2018+107+118}, \cite{schreier2022Data}. One prominent environment representation are \textit{Dynamic Grid Maps} (DGMs) \cite{steyer2018grid}, in which the vehicle-centered environment in Bird's-Eye-View (BEV) is divided into grid cells, each of which is assigned a probability of being occupied or free as well as a velocity distribution. 
	The dynamic grid map enables tasks such as free space extraction \cite{Schreier16Compact} and detection of arbitrary-shaped objects.\\ 
        \begin{figure}[t]
		\centering
            {\setlength{\fboxsep}{0pt}%
            \setlength{\fboxrule}{0pt}%
		\framebox{\includegraphics[width=0.485\textwidth]{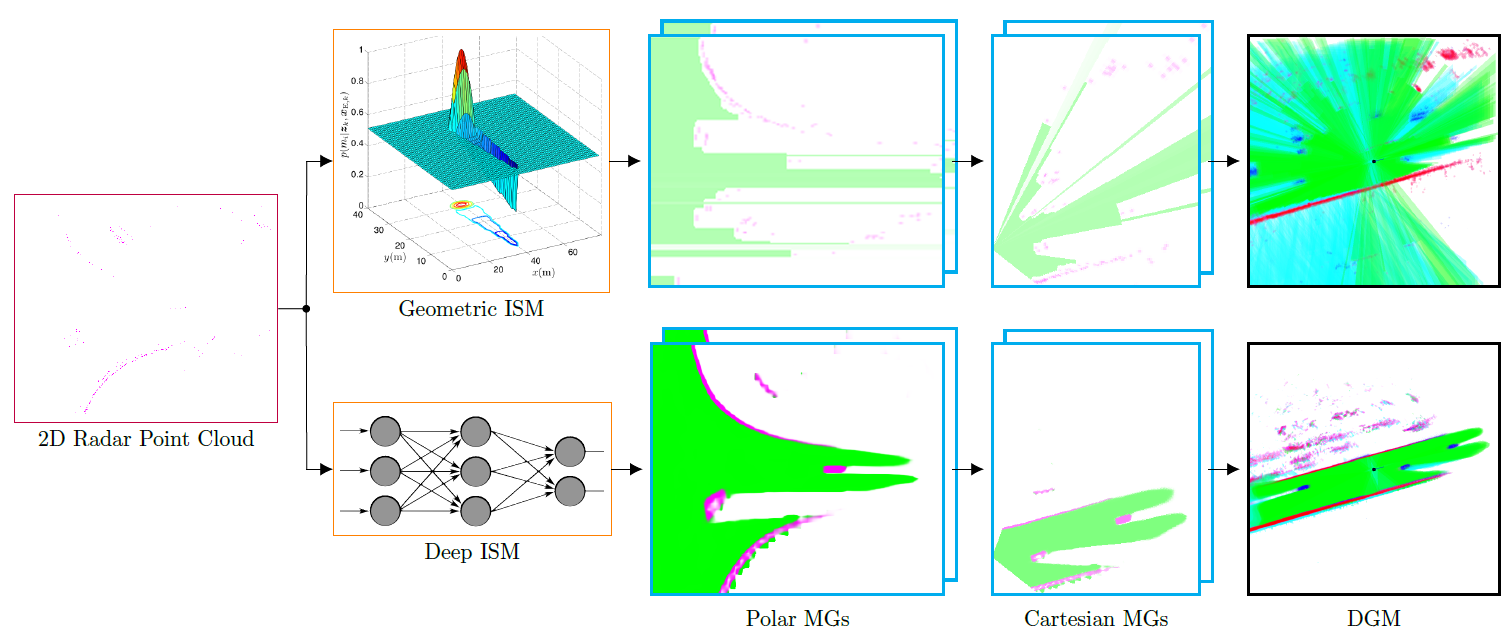}}}
		\caption{Overview of geometric ISMs (top) and deep ISMs (bottom). For illustrative purposes, the 2D RADAR point cloud is colored in pink. A front RADAR is chosen to show the exemplary Measurement Grid (MG). The MGs are then used as inputs for a dynamic grid fusion algorithm to estimate a DGM, which jointly represents the static and dynamic driving environment in a consistent manner. The ego-vehicle is denoted as a black dot with a directional bar in the center of the DGMs, free areas are shown in green, statically occupied areas in red, and dynamically occupied areas in blue.}
		\label{figure:framework}
	\end{figure}
	To generate such DGMs, one prerequisite is to design Inverse Sensor Models (ISMs) to generate \textit{measurement grids} given the sensor measurements. One favorable sensor choice is LiDAR because of its dense and accurate range detections. Compared to LiDAR, RADAR detections have a longer range, a direct velocity measurement and can operate in severe weather conditions. However, since RADAR detections from most automotive RADARs are sparse and uncertain (e.g. due to multipath reflections), it is difficult to correctly infer occupied and free areas of the surrounding environment purely from individual RADAR measurements. Until now, hand-designed geometric ISMs with many parameters are often in use to generate RADAR measurement grids. Accumulating RADAR measurement grids during the generation of DGMs can attain a satisfactory level of certainty about the environment. However, the problem becomes harder in high-speed scenarios, where the accumulation time for the same scene is limited. The suboptimal estimation results and tedious parameter optimization further motivate us to apply a deep ISM.\\In this work, we handle the sparsity problem in highway driving scenarios via a neural network by modeling the problem of learning an ISM as a 2D image semantic segmentation task. 
We directly train our network, the Dual Attention Network (DANet) \cite{fu2019dual}, in polar coordinates with LiDAR-based annotation of free/occupied cells. In the inference stage, for each RADAR, a single-frame sparse scan is firstly discretized into a 2D polar map as the input of the semantic segmentation framework. As a result, the network outputs a measurement grid containing a distribution over different states (occupied, free and unknown) for each cell. The learned measurement grids are combined with RADAR Doppler velocity measurements, converted into Cartesian coordinates and then used to generate DGMs. An overview of geometric ISMs vs. the proposed deep ISMs for generating DGMs is shown in Fig. \ref{figure:framework}.\\To summarize, our contributions are:
	\begin{itemize}
		\item We implemented a deep learning framework to learn a complex RADAR ISM focusing on challenging real-world highway driving scenarios. 
		\item Compared to the other deep ISMs, we train the network in polar coordinates for single frames of single RADARs, as opposed to collecting (accumulated) points from multiple RADARs to cover 360° in Cartesian coordinates. Learning in the local polar scheme makes our ISM independent of sensor mounting.
		\item Compared to the other deep ISMs, we are the first to generate RADAR-based DGMs by combining the estimation of deep ISM with Doppler measurements from the RADAR. 
		\item The structure of our framework does not limit the number of applied RADARs. It can also flexibly integrate other sensor modalities (such as camera and LiDAR) with either geometric or deep ISM.
	\end{itemize}
	\section{RELATED WORK}
	A discrete 2D BEV measurement grid $\textbf{g}$ can be described either probabilistically or evidentially \cite{dempster1968generalization}, \cite{shafer1976mathematical}. The former represents each grid cell as a discrete probability distribution over free and occupied states in (\ref{eq_pro}), whereas the latter assigns the grid cell Dempster-Shafer belief masses of being free, occupied or unknown $b_F, b_O, b_{FO} \in[0,1]$ in (\ref{eq2}). 
	\begin{align}
		\textbf{g}^{(pro)} &= [p_F, p_O]^{T} \label{eq_pro}\\
		\textbf{g}^{(evi)} &= [b_F, b_O, b_{FO}]^{T}\label{eq2}
	\end{align} 
	Our approach uses the latter since the unknown mass is explicitly considered. The measurement grid state $\textbf{g}^{(evi)}$ can be estimated by an ISM given the sensor measurement. The measurement grids are later fused over several time steps to generate a dynamic occupancy grid map. ISMs can be categorized as either hand-crafted geometric ISMs or deep learning-based ISMs.
	\subsection{Geometric ISMs}
	A geometric ISM \cite{2021:steyer:gridbased}, \cite{thrun2006probabilistic}, \cite{7117922} exploits geometric information inherent in the measurement with hand-crafted models to calculate a mapping from sensor measurements to occupancy masses and free masses in the environment. These algorithms map
	a sensor detection point to one or more occupied cells in consideration of the sensor uncertainties. The cells
	between the sensor and the occupied cells are often implicitly estimated
	as free. Geometric ISMs are well-suited for 3D LiDAR point clouds, as the data is dense enough to estimate meaningful grid states. 
	Compared to LiDAR, RADAR scans are noisier and sparser, resulting in false positives and an incomplete representation of the environment. 
	Moreover, geometric RADAR ISMs require a large number of parameters to adequately consider sensor uncertainties, making parameter optimization difficult. Additionally, these methods only consider the geometric correlation between the sensor position and detection, neglecting the spatial coherence of RADAR detections.
	\subsection{Deep ISMs}
	Compared to geometric ISMs, deep learning-based ISMs take into account the spatial coherence of all detection points, thus considering the scene context. In \cite{bauer2019deep_evidential}, sensor scans from all RADARs are collected to cover 360° around vehicle and then discretized into cells in Cartesian coordinates to feed into the deep ISM. \cite{bauer2020prior} combines geometric ISM with deep ISM in \cite{bauer2019deep_evidential} to generate occupancy grid maps. \cite{Weston19unknown} uses point cloud in polar coordinates from a special RADAR (with 360° FOV but without Doppler) as input and train a deep ISM to get measurement grids in Cartesian coordinates. 
	In \cite{sless2019road}, RADAR scans are discretized into cells in Cartesian coordinates and used as input for the network. However, this method requires masking out the region outside the FOV in both input and ground truth data, filtering out dynamic RADAR points with Doppler velocity measurements, and focusing only on static detections. Additionally, the study investigates the effect of aggregating RADAR frames, but such aggregations are hard to perform in highly dynamic driving scenarios in which temporally consecutive RADAR point clouds differ strongly.\\
	To our knowledge, our approach is the first to learn the measurement grid state in a single polar frame from RADARs with limited FOV, which makes it independent of sensor mounting positions and eliminates the need to mask out the FOV. This allows our ISM to be easily applied to new RADARs of the same type. Additionally, unlike \cite{sless2019road}, \cite{bauer2019deep_probability}, we incorporate dynamic RADAR detections and use Doppler measurements to estimate the cell velocity during the generation of a DGM. Moreover, our work is the first one to couple the deep ISM with a dynamic grid fusion algorithm.
	\subsection{Dynamic Grid Maps (DGMs)}
	DGMs are an extension to the well-known occupancy grid maps by adding cell velocity information. Therefore, it is more suitable for dynamic environments in autonomous driving. The velocity is normally estimated by particle-based algorithms \cite{steyer2018grid}, \cite{6907756}, \cite{tanzmeister2016evidential}, \cite{negre2014hybrid}, \cite{nuss2018random}, in which particles are assigned to cells. In a first step, the particles are predicted and free to move between cells. In a second step, the predicted DGM is updated with the current measurement grid. For RADARs, the newborn particles’ velocity can be initialized with Doppler velocity measurements. In our work, the method described in \cite{steyer2018grid} is used for generating DGMs, where the occupied cells are further categorized into static/dynamic to better represent static obstacles, dynamic objects, and free space.
	\section{Method}
	\subsection{Evidential Occupancy Representation}
	\label{sec:3.1}
	We follow the idea of \cite{steyer2018grid} to use the Dempster-Shafer Theory (DST) of evidence to represent different dynamic grid cell states:
	\begin{align}
		\Theta &= \{F, D, S\},\\
		2^{\Theta} &= \{\emptyset, \{F\}, \{D\}, \{S\}, \{F, D\}, \{D, S\}, \{F, S\}, \Theta\}.
	\end{align}
	The case when a cell is both free and static is excluded: $\{F, S\} \coloneqq \emptyset$. 
	In the DST Framework, a mass $m(\cdot)$ is assigned to each hypothesis of the power set $2^{\Theta}$ such that a unit sum is attained. 
	Similarly, we use Dempster-Shafer masses to represent grid cell states in a measurement grid:
	\begin{align}
		\Theta &= \{F, O\},\\
		2^{\Theta} &= \{\emptyset, \{F\}, \{O\},\Theta\}.
	\end{align}
	A mass $b(\cdot)$ is assigned to each hypothesis. For visualization of DGMs and measurement grids in this work, we select the same color coding scheme as in \cite{steyer2018grid}:
	\begin{align}
		RGB = (1- \sum_{{S}\cap \Theta=\emptyset}{m_{\theta}}, 
		1- \sum_{{F}\cap \Theta=\emptyset}{m_{\theta}},
		1- \sum_{{D}\cap \Theta=\emptyset}{m_{\theta}})\label{eq:6}
	\end{align}
	\subsection{Generation of RADAR Input Data}
	\label{sec:model:generation_of_RADAR_input_data}
	Instead of learning measurement grids from scans of multiple RADARs (\cite{bauer2019deep_evidential}, \cite{bauer2020prior}), we propose to learn the occupancy state in the local sensor scheme. We define the input as a 2D polar map of azimuth and range bins $A\times R$ with bin sizes of $\alpha_{A}$ in° per azimuth bin and $\alpha_{R}$ in meters per range bin. This map covers the RADAR-centered FOV: $0$ to $\alpha_{R}R$ in range and $-0.5 \alpha_{A}A$ to $0.5 \alpha_{A}A$ in azimuth. The RADAR detections are discretized into a binary image, with 1 meaning at least one valid RADAR detection in that cell. 
	\subsection{Generation of LiDAR-based Labels}
	\label{sec:model:generation_of_LiDAR_based_label}
	In order to generate our ground truth data, we generate LiDAR-based DGMs at first. A ray tracing algorithm takes one dense LiDAR point cloud as input to generate a single-frame LiDAR measurement grid. Then the iterative prediction-update algorithm in \cite{steyer2018grid} fuses multiple LiDAR measurement grids over several time steps to generate a LiDAR-based DGM. Given a DGM, we convert it to the same form as a measurement grid by calculating the free, occupied and unknown belief masses $b_{F}$, $b_{O}$, $b_{FO}$:
	\begin{align}
		&b_O=m_D+m_S+m_{DS},\\
		&b_F=m_F,\\
		&b_{FO}=1 - b_{F}-b_{O} =:u.
	\end{align}
Compared to a single-frame LiDAR measurement grid, this accumulated data is more informative and can avoid temporary occlusion, and thus serves as a high-quality, fully auto-labeled ground truth.\\
	\begin{figure}[t]
		\centering
            {\setlength{\fboxsep}{0pt}%
            \setlength{\fboxrule}{0pt}%
		\framebox{\includegraphics[width=0.5\textwidth]{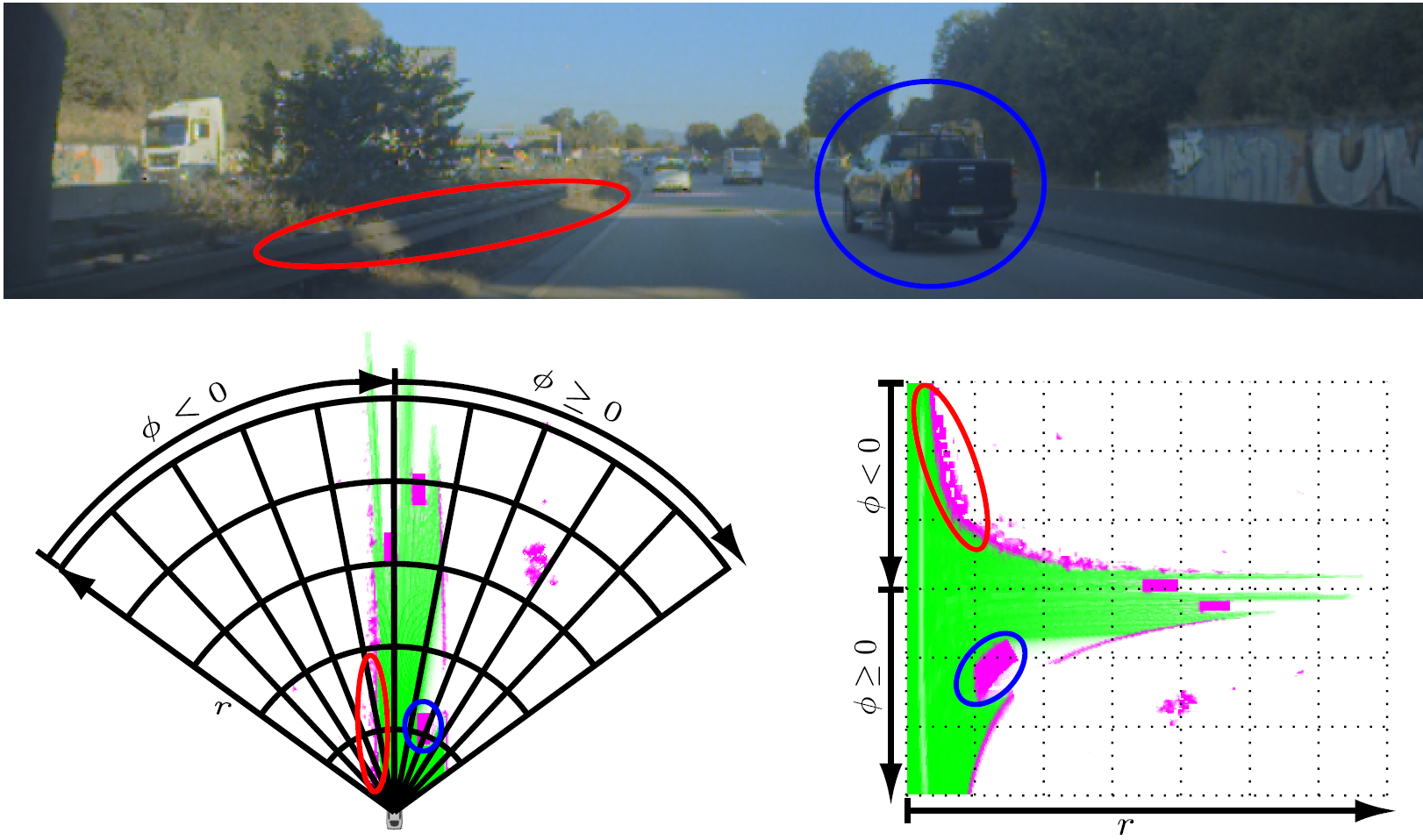}}}
		\caption{Exemplary illustration of the involved coordinate conversion on the front RADAR in a highway scenario.\\a) Front camera image. b) $\textbf{g}^{(evi)}$ of the area related to the front RADAR in Cartesian coordinates. c) $\textbf{g}^{(evi)}$ in polar coordinates.\\
        A guardrail (red) and a vehicle (blue) are exemplarily shown as corresponding ellipses in all images. $r$: range, $\phi$: azimuth angle.}
		\label{figure:3.1}
	\end{figure}
Moreover, two improvements are performed to the LiDAR-based labels. First, since dynamic objects appear most of the time in L-shape in a DGM, the dynamic objects extracted as oriented bounding boxes from LiDAR-based DGMs (see Chapter 4 in \cite{2021:steyer:gridbased}) are used in the labeling. For the area of these oriented bounding boxes in a LiDAR ``measurement grid'', $b_O$ is set as 1. This step refines the object shape and thus improves the inference results of dynamic objects.\\
	Then, the occupancy information related to a certain RADAR is preserved and transformed into a polar map of size and precision $(A, R, \alpha_{A}, \alpha_{R})$ as shown in Fig. \ref{figure:3.1}. In the polar coordinates, the second improvement is performed in the partially observed area, which is between the first and last return along a ray. In this area, we exclude the irrelevant occupancy information $b_O$ where no RADAR measurement points are present. A morphological dilation is applied to preserve $b_O$ around the RADAR detections. This will prevent the neural network from creating unmeaningful structures in the partially observed area, which is normally outside the road. Finally, one of the classes \textit{free}, \textit{occupied}, or \textit{unknown} is assigned for each cell. The resulting 2D matrix of size $A\times R$ is used as annotation data.
	\subsection{Model}
	\subsubsection{Network architecture} 
	Following the idea of \cite{sless2019road}, we treat the occupancy inference problem as a 3-class 2D image semantic segmentation task. Self-attention mechanism is chosen for better considering the spatial relationship between RADAR detection points. \\
	The network architecture of our framework is based on the popular Dual Attention Network (DANet) \cite{fu2019dual}, which captures the semantic interdependence in spatial and channel dimensions by two attention modules with relatively low execution time. This network is divided into two parts. First, the backbone ResNet extracts low-level features from the sparse 2D RADAR point cloud and encodes them into denser features. Then the decoder head achieves pixel-level belief estimation with the position attention module and channel attention module. The position attention module performs a weighted sum of features from all positions to selectively aggregate the feature at each position. This considers related features regardless of their distances. The channel attention module integrates associated features among all channel maps to selectively emphasize interdependent channel maps. As a result, the summation of the outputs from both attention modules enhances the feature representation, leading to improved occupancy belief assignment with greater precision. Furthermore, to incorporate the learned beliefs into the DGM, inspired by SoftNet \cite{bauer2019deep_evidential}, we retrieve the occupancy evidence and free evidence from the output of the softmax layer in our segmentation network.
	\subsubsection{Loss}
	Similar to other works \cite{sless2019road}, \cite{bauer2019deep_probability}, severe class imbalance problem is present in our dataset, where only about $2\%$ pixels in annotation images belong to the class \textit{occupied}. According to \cite{jadon2020survey}, two categories of loss can be used to address class imbalance: region-based loss (e.g., Dice Loss, Lavosz Loss) and distribution-based loss (e.g., Focal Loss, weighted Cross Entropy). In the field of deep ISM, \cite{sless2019road} uses region-based Lavosz Loss to explicitly solve the class imbalance problem. \cite{bauer2019deep_probability}, \cite{Senanayake2017DeepOM} trained a neural autoencoder with weighted mean squared error with two weighting schemes (1. Inverse Class Ratio Weighting Scheme 2. Independent Class Weighting Scheme). They further showed that with the former strategy, the measurement grid state $\textbf{g}^{(pro)}$ is better inferred.\\ We found Dice Loss to perform best in our proposed method. Minimizing dice loss in (\ref{eq:11}) is equivalent to maximizing the dice coefficient (F1 score), where $I$ represents the intersection of correct estimation and label, $U$ represents the union of estimation and label, and $\epsilon$ is a smoothing term to smooth loss and gradient. 
	\begin{align}
		L_{dice}=\frac{2I+\epsilon}{U+\epsilon} \label{eq:11}
	\end{align}
	
	\section{EXPERIMENTAL SETUP}
	\begin{figure}[b]
		\centering
		\includegraphics[width=0.3\textwidth]{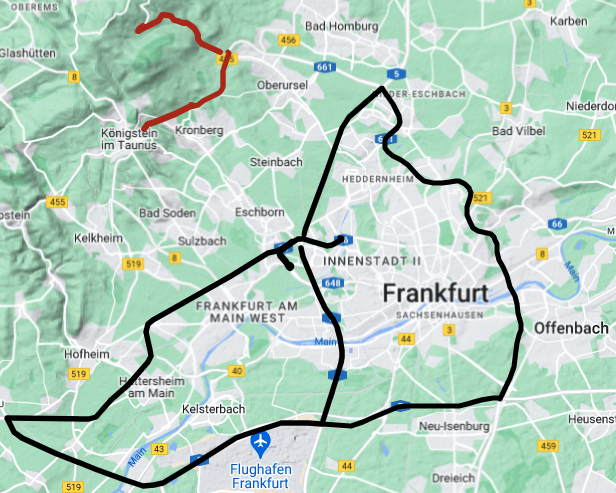}
		\caption{Driven routes for dataset generation: test set (red), training and validation set (black)}
		\label{figure:4.1}
	\end{figure}
	\subsection{Baseline: RADAR Geometric ISM}
	\label{sec:baseline}
	The frequently used geometric ISM in \cite{2021:steyer:gridbased} was implemented as reference. It models the occupancy evidence by a mixture of 2-D Gaussian distributions of individual measurements, with Gaussian uncertainty distributions in angular and radial direction. In this process, the Doppler velocity measurement of each detection point is compensated by the ego movement, preserved in the corresponding cell and spread to the neighboring cells. This absolute radial velocity is used to improve the dynamic estimation during the generation of DGMs. The free mass is implicitly derived between the sensor origin and the first return along the ray. To deal with false positives, the parameters of occupancy and free space evidences have to be selected conservatively. That’s why the free mass and occupancy mass are small in the measurement grid from a geometric ISM, as shown in Fig. \ref{figure:framework} and \ref{fig:5.3}. Based on experiments, the size of the output measurement grid for geometric baseline is manually optimized as $A = 108$ (azimuth bins), $R = 200$ (range bins) with bin sizes of $\alpha_{A} = 1$° per azimuth bin and $\alpha_{R}= 0.5\, m$ per range bin.
	\subsection{Data Collection}
	\label{sec:data collection}
	The sensor data is collected from real-world highway driving scenarios (including highway entrance and exit). The vehicle is equipped with six Continental SRR 520 RADARs (four mounted at the corners, one central at the front and one central at the rear) and one roof-mounted VLS-128 LiDAR. The nominal update rate of VLS-128 is $10\,Hz$. To reduce the amount of similar training data, only every 8-th DGM
is used to generate the lidar annotation data. To each reference DGM, the nearest frame of each RADAR is found and ego-motion compensation is performed. In this way, each DGM is converted to 6 polar measurement grid images (see Sec. \ref{sec:model:generation_of_LiDAR_based_label}).
	We choose different locations to collect training and test data, as depicted in Figure \ref{figure:4.1}. The vehicle drove in both directions of the high-speed roads along the routes. The total driving route is ca. $180\,km$ for training (56,568 images) and validation (16,530 images), and ca. $30\,km$ for test (12,180 images).
	\subsection{Implementation Details}
	\label{sec:implementation details}
	The cell sizes of measurement grids in Cartesian coordinates and DGMs are both set as $\alpha_{x} = \alpha_{y} = 0.2\,m$ in this work, where the dimension of both square grids is chosen as $700 \times 700$. 
	During training, the data is flipped with a 50\% of chance in the azimuth direction. The sizes of the LiDAR labels, the input image and the output measurement grid of the network are all set as A = 300 (azimuth bins), R = 350 (range bins), with azimuth bin size as $\alpha_{A}$ = 0.36° and range bin size as $\alpha_{R}=0.2\,m$. The bin sizes are chosen so small that a minimum information loss of LiDAR reference during the coordinate transform from Cartesian to polar scheme is attained (especially for cells with a longer range). 
	The network is trained with a momentum-based SGD optimizer with an initial learning rate of $0.003$. In the decoder head, a dropout layer with probability 0.1 is applied to avoid overfitting.
	\section{Results and Discussion}
	\label{sec:5}
	In the first subsection, we will verify one of the most important advantages of our approach, i.e., the independence of the RADAR mounting position. For this purpose, the outputs
of the neural network for different RADAR mounting positions will be compared quantitatively and qualitatively. Then, the effect of pre-aggregation on the performance of our deep ISM is shown in Sec. \ref{sec:5.2}. In Sec.
	\ref{sec:5.3}, our approach is compared with the baseline method – a geometric ISM. In the end, the qualitative comparison of RADAR DGMs generated by our deep ISM and by the geometric ISM is presented in Sec. \ref{sec:5.4}.
	\subsection{Deep ISM: Effect of RADAR Mounting}
	\label{sec:5.1}
	\begin{figure}[t]
		\centering
            {\setlength{\fboxsep}{0pt}%
            \setlength{\fboxrule}{0pt}%
		\framebox{\includegraphics[width=0.47\textwidth]{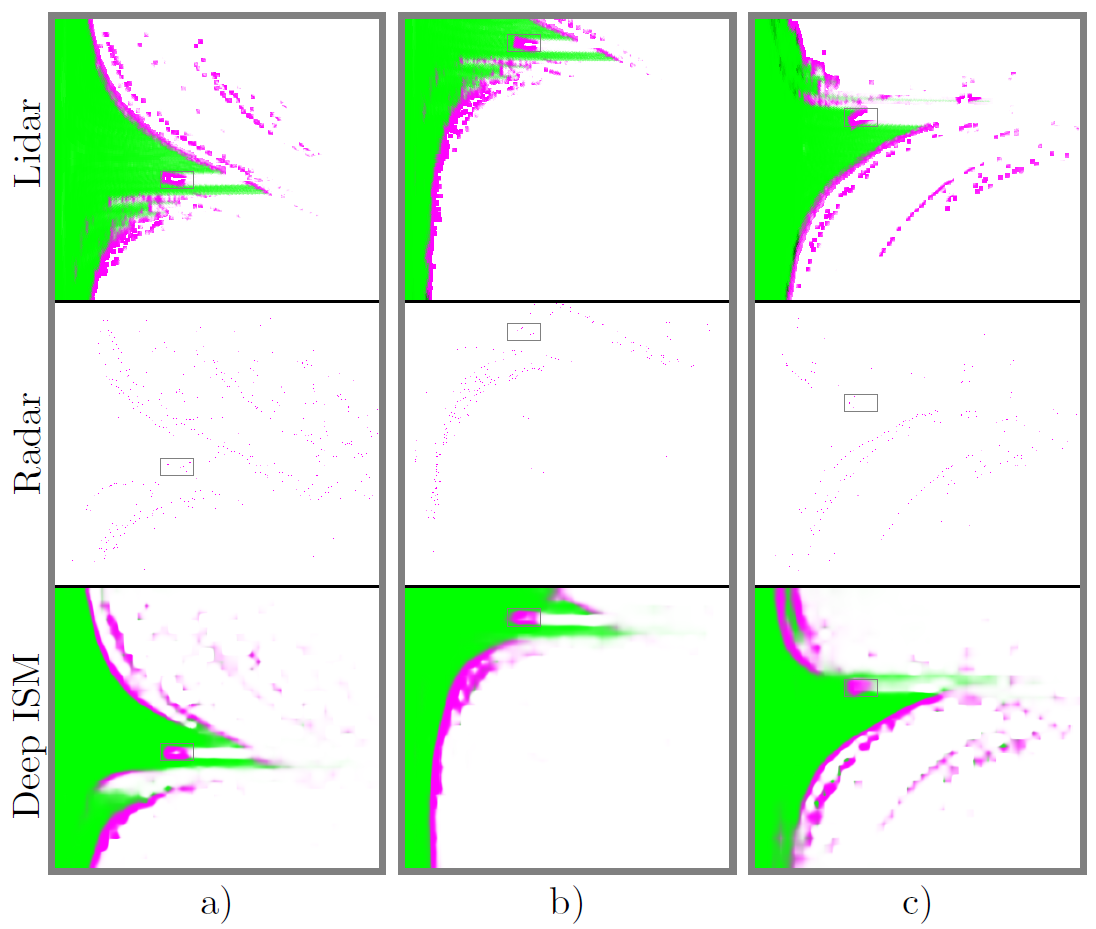}}}
		\caption{Qualitative result on setup B with different RADARs, here a) front center RADAR, b) front left RADAR, c) rear center RADAR. Rows from top to bottom are: filtered LiDAR-based annotation before thresholding, RADAR detections, RADAR measurement grid. The dynamic object is indicated manually with a bounding box.}
		\label{fig:5.1}
	\end{figure}
	\begin{table}[b]
		\caption{Mounting Differences Between The Two Setups A and B ($\Delta_X = X_B - X_A$)}
		\label{table:1}
		\begin{center}
			\begin{tabular}{l||c|c|c|c}
				\hline
				Sensor & $\Delta_{x}[m]$ & $\Delta_{y}[m]$ & $\Delta_{z}[m]$ &$\Delta_{yaw}$[$\deg$]\\
				\hline
				front center &-0.0780 &-0.1260 &0.3660 &0.2167 \\
				front left &-0.0230 &-0.0267 &-0.5160 &-0.0661\\
				rear center &-0.0384 &1.0111 &-0.5586 &-0.5267
			\end{tabular}
		\end{center}
	\end{table}
	\begin{table}[bt]
		\caption{Class IoU Comparison for the Front Center, Rear Center and Front Left RADAR of 2 Setups}
		\label{table:2}
		\begin{center}
			\begin{tabular}{l||c|c|c|c}
				\hline
				Sensor & mIoU.fr & mIoU.oc & mIoU.un &mIoU\\
				\hline
				B.front center& 77.22 & 23.83 &89.51 &63.52\\
				A.front center & 84.05 & 31.22 &90.80 &68.70\\
				\hline
				B.rear center & 76.32& \textbf{24.61}& 87.54& 62.82\\
				A.rear center &82.50 & 22.40 & 88.50 &64.52 \\
				\hline
				B.front left &77.40&\textbf{28.80}&91.64&65.95\\
				A.front left &77.90 &26.40 &94.40 &66.20\\
				\hline
			\end{tabular}
		\end{center}
	\end{table}
	To evaluate the effect of RADAR mounting, we test the result of the neural network, i.e., the learned measurement grids in polar coordinates. The network is trained for all the six single RADARs in setup A.
	We further exploit another setup B, also with 6 RADARs only for testing. Each RADAR in setup B has a slightly different mounting position compared to setup A. For brevity, we restrict our further discussion to the front left corner RADAR as 4 corner RADARs are mounted symmetrically and they show similar results in this experiment. The mounting differences of the front center RADAR, the rear center RADAR and the front left RADAR are shown in Table \ref{table:1}.\\
	Table \ref{table:2} presents that the performance on setup B did not show a large decrease compared to the inference results of setup A. It’s notable that the per class mean intersection over union (mIoU) scores of occupied (mIoU.oc) is generally much lower than other IoU values. This is because the RADAR point cloud from SRR520 RADARs is more sparse than the LiDAR point cloud from the VLS-128. For rear center and front left RADAR, the mIoU.oc is even higher for a different installation position.\\The qualitative results of setup B are shown in Fig. \ref{fig:5.1}. Although the data from setup B was not used for training, the vehicles, guardrails and free spaces are all correctly represented as expected. The experiment verifies that the trained network can be applied to a new RADAR of the same type independent of its mounting.
	\subsection{Deep ISM: Effect of Time Frame Aggregation}
	\label{sec:5.2}
	\begin{figure}[b]
		\centering
            {\setlength{\fboxsep}{0pt}%
            \setlength{\fboxrule}{0pt}%
		\framebox{\includegraphics[width=0.47\textwidth]{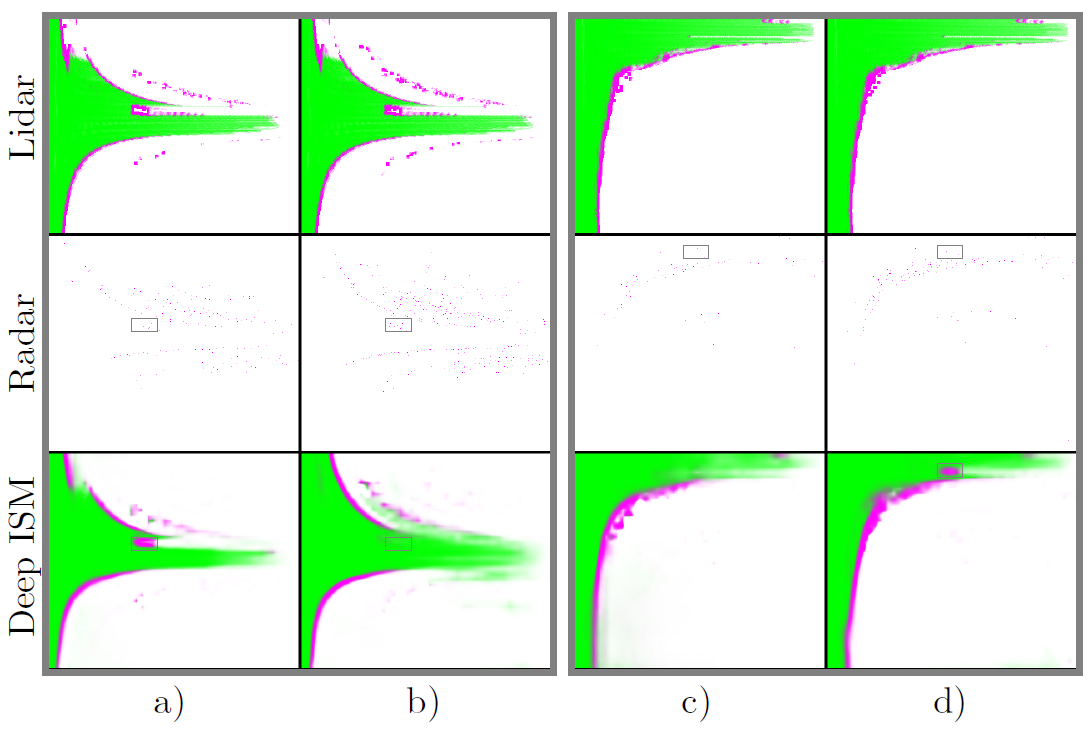}}}
		\caption{Polar measurement grid comparison between single frame and double frame. Here different RADARs are shown: a) front center, single frame. b) front center, double frames. c) front left, single frame. d) front left, double frames. The dynamic object is indicated with a bounding box. Rows from top to bottom are: filtered LiDAR-based annotation before thresholding, RADAR detections, RADAR measurement grid.}
		\label{fig:5.2}
	\end{figure}
	To deal with the sparsity of RADAR measurements, many works, e.g. \cite{sless2019road}, have applied time frame aggregation with ego vehicle odometry compensation. It's interesting to see if the pre-aggregation can improve our deep ISM. Since the aggregation is based on the assumption that all the cells are static, it would cause the loss of dynamic occupancy belief estimation, especially in high-speed scenarios. Given the update rate $\alpha_{t}$ of a single SRR 520 RADAR (about $20\,Hz$), the speed $v$ of a dynamic object (the average speed of passenger vehicles on German highways is $125\,$km/h) and the number of aggregated frames $n$,  then the introduced spatial error $\delta_{L} = \alpha_{t} vn$ should be at least less than the vehicle length (the average length of a passenger vehicle is around $4.5\,m$). Therefore, in our work $n$ is set to 2 for an acceptable error for highway driving. \\Results in Table \ref{table:3} show that for our deep ISM, aggregation over two RADAR frames does not contribute to a significant performance boost. Moreover, although the mIoU of class occupied is slightly improved, both object related RADAR points and noise points increase. Therefore more RADAR points are required to be detected as an object. More specifically, comparing the results in Fig. \ref{fig:5.2} a) and in Fig. \ref{fig:5.2} b), the deep ISM takes the aggregated object-related RADAR points as sensor noise and causes false negative occupied belief (denoted by a bounding box). d) shows an example of false positive estimates due to occasionally accumulated noise-related RADAR detections inside a certain region. The conclusion can be drawn that the training with aggregation is not beneficial in our tested scenarios. Therefore, the later experiments will be carried out for single frame RADAR measurement grids.\\
Instead of pre-aggregation, our approach accumulates measurement grids only in the generation process of DGMs. This was found superior than the pre-aggregation of RADAR frames under the assumption of only static cells.  
	\begin{table}[t]
		\caption{Class IoU under different aggregation}
		\label{table:3}
		\begin{center}
			\begin{tabular}{c||c|c|c|c}
				\hline
				Methods & mIoU.fr & mIoU.oc & mIoU.un &mIoU\\
				\hline
				1 Frame ISM & 79.98 & 27.4 & 94.91 &65.77\\
				2 Frame ISM & 79.10 & 30.08 & 93.44 &67.54\\
				\hline
			\end{tabular}
		\end{center}
	\end{table}
	\subsection{Deep ISM: Comparison to the Geometric ISM}
	\label{sec:5.3}
\begin{figure}[h]
		\centering
		{\setlength{\fboxsep}{0pt}%
			\setlength{\fboxrule}{0pt}%
			\framebox{\includegraphics[width=0.49\textwidth]{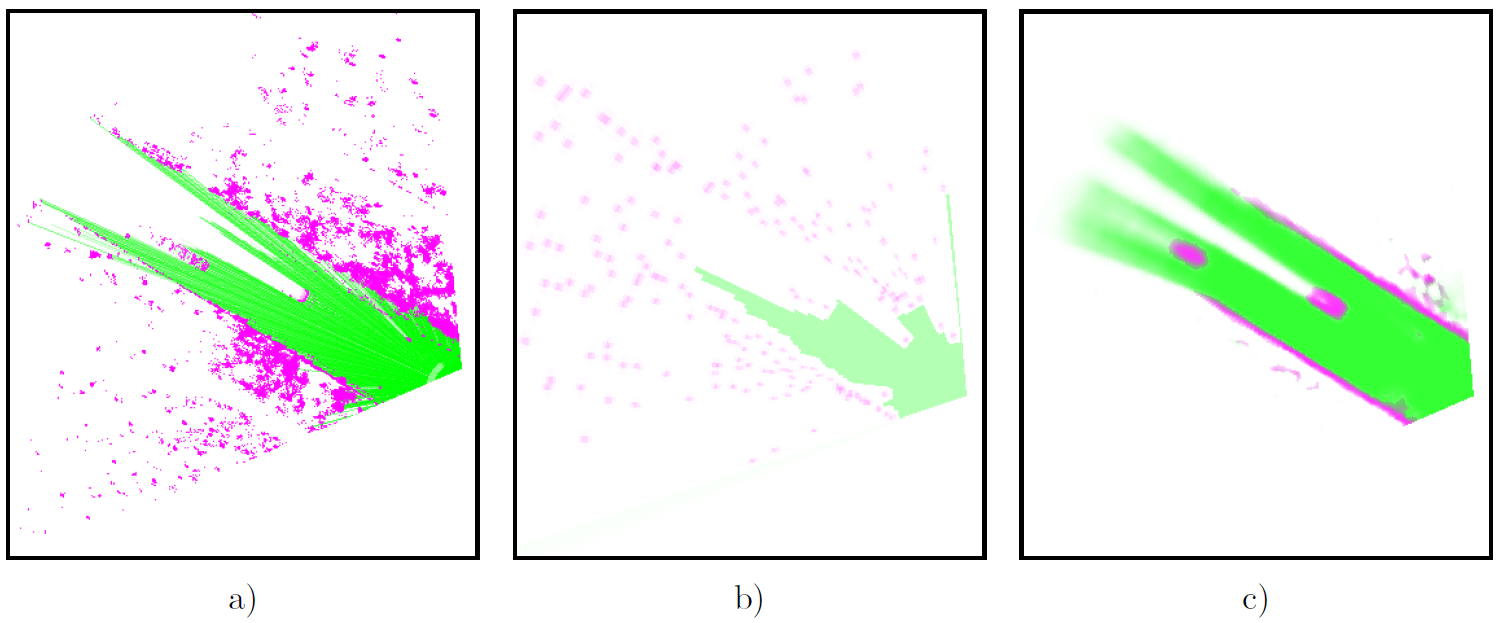}}}
		\caption{Qualitative result comparison in vehicle centered measurement grids (Cartesian) of deep ISM and geometric ISM for the front RADAR. a) LiDAR \enquote{measurement grid} without improvements (see Sec. \ref{sec:model:generation_of_LiDAR_based_label}). b) Geometric RADAR measurement grid. c) Deep RADAR measurement grid.}
		\label{fig:5.3}
	\end{figure}

	Since measurement grids only in polar coordinates from geometric ISMs and from deep ISMs have different sizes and resolutions, we decide to compare measurement grids in Cartesian coordinates. The conditional probabilities of the state estimates given reference states, inspired by \cite{bauer2019deep_evidential}, are applied as the evaluation metric.\\
Results comparing the performance between our approach and the geometric ISM baseline are presented in Table \ref{table:5.3}. It shows that our deep ISM outperforms the classic approach using Gaussian Mixture Models by a large margin. We show one example of the front sensor in this case in Fig. \ref{fig:5.3}. Here, the occupancy map state estimation is shown in Cartesian coordinates for an ego-vehicle-centered grid map. In spite of pre-filtering of detections with RADAR cross section (RCS) less than $-20\,dB/m^2$ in the baseline method, the geometric ISM is still prone to noisy detections. The results show that our approach eliminates the effect of noisy RADAR detection points, and learns a complex ISM to predict denser guardrails and dynamic objects.
\begin{table}[h]
		\caption{Conditional Probabilities of the
Estimates $\hat{(\cdot)}$ given Reference States $\Tilde{(\cdot)}$ for Deep ISM and Geometric ISM}
		\label{table:5.3}
		\begin{center}
			\begin{tabular}{c|c|c|c||c|c|c}
				\hline
				Methods & $p_{(\hat{F}|\Tilde{F})}$ & $p_{(\hat{O}|\Tilde{F})}$ & $p_{(\hat{u}|\Tilde{F})}$ & $p_{(\hat{F}|\Tilde{O})}$ & $p_{(\hat{O}|\Tilde{O})}$ &$p_{(\hat{u}|\Tilde{O})}$\\
				\hline
				Baseline & 60.26 & 3.71 & 36.13 &48.55 &15.30&36.13\\
				
				Deep ISM & 85.74 &3.20	&11.05 &19.92 &38.68 &41.40\\
				\hline
			\end{tabular}
		\end{center}
	\end{table}	
\subsection{Deep ISMs for Dynamic Grid Mapping: RADAR DGM}
	\label{sec:5.4}
\begin{figure}[t]
		\centering
            {\setlength{\fboxsep}{0pt}%
			\setlength{\fboxrule}{0pt}%
			\framebox{\includegraphics[width=0.49\textwidth]{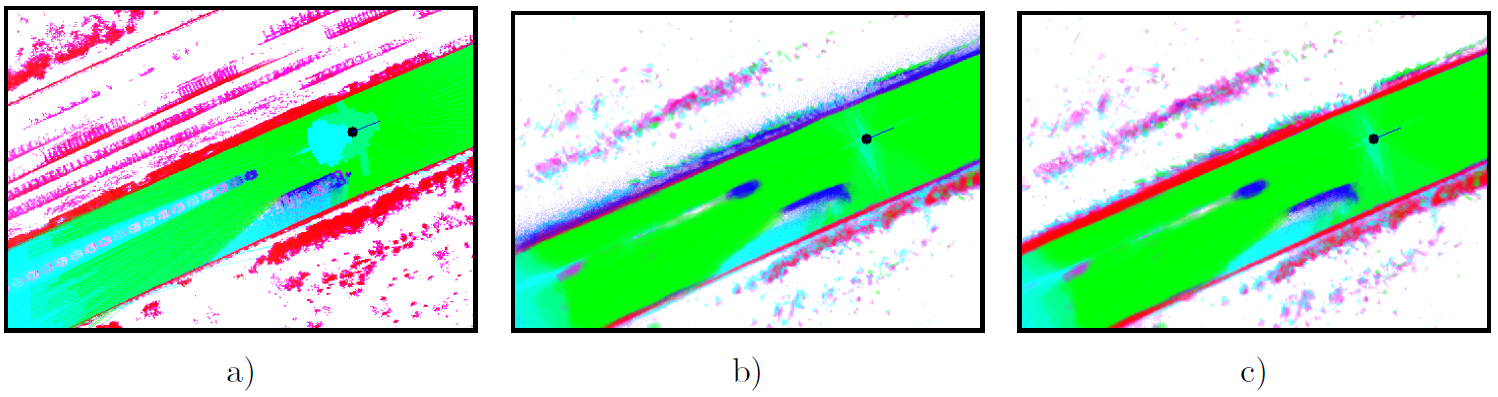}}}
		\caption{Effect of adding Doppler information. a) LiDAR reference. b) deep RADAR DGM without RADAR Doppler information c) deep RADAR DGM with RADAR Doppler information. Here the ego-vehicle is depicted with a black dot with orientation bar.}
		\label{fig:doppler}
	\end{figure}

\begin{figure*}[t]
		\centering
            {\setlength{\fboxsep}{0pt}%
			\setlength{\fboxrule}{0pt}%
			\framebox{\includegraphics[width=1\textwidth]{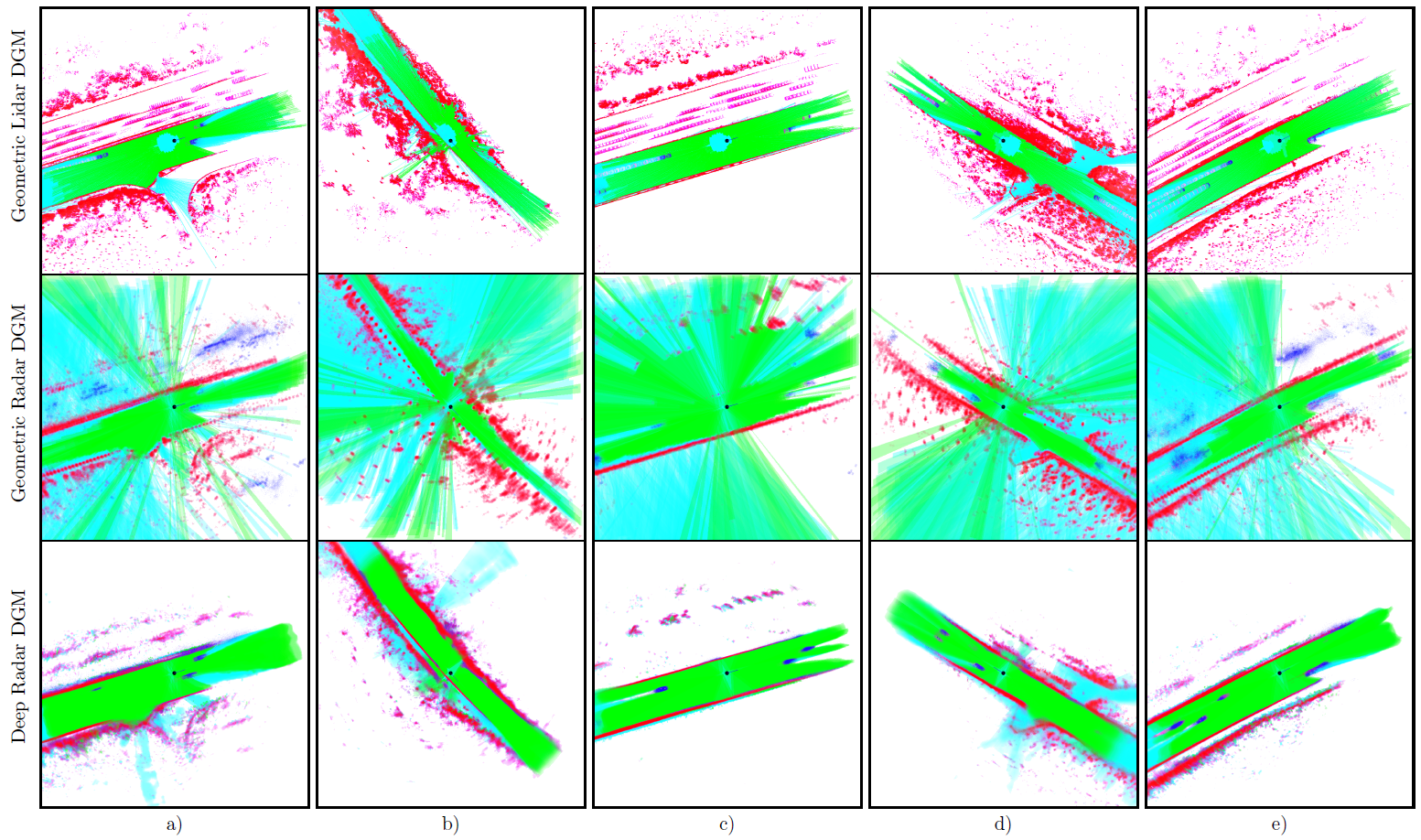}}}
		\caption{Qualitative Result of RADAR DGMs. Each column represents a different scene. Rows from top to bottom are: geometric LiDAR DGMs, geometric RADAR DGMs, deep RADAR DGMs. In all cases, the deep RADAR DGMs show closer similarity to the high-fidelity LiDAR DGMs than the Geometric RADAR DGMs. a), c), d), e): Cells occupied by dynamic objects are better detected with deep ISM; b), c): Guardrails are better represented with deep ISM.}
		\label{fig:5.4}
	\end{figure*}

	The resulting RADAR measurement grids based on our deep ISM are further used as inputs for the generation of DGMs.
	Similar as in the baseline (Sec. \ref{sec:baseline}), the compensated Doppler velocity measurement, i.e., the absolute radial velocity, is used for a better dynamic estimation during the generation of DGMs.
	The velocity information of each RADAR point is copied to the corresponding cell and its eight direct neighboring cells in the learned measurement grid, when the cell’s occupancy mass is larger than a threshold of $b_O \geq 0.196$ (set experimentally). Blue cells on the guardrail in Fig. \ref{fig:doppler} b) shows that the DGM without Doppler information has a lot of false positive dynamic cells.\\
	The further steps from measurement grid to DGM are identical for both deep ISM and geometric ISM. The measurement grid in polar coordinates is converted to Cartesian coordinates and used in the DGM estimation process (see Fig. \ref{figure:framework}). For RADARs, newborn particles’ velocities are initialized with compensated Doppler velocity measurements.\\
	Fig. \ref{fig:5.4} shows the comparison of RADAR DGMs created via deep ISM and via geometric ISM in five different scenes.  
 In all the scenes, especially in Fig. \ref{fig:5.4} c), guardrails are better represented with deep ISM than with the baseline model. Regarding dynamic estimates (blue), thanks to Doppler velocity measurements and improved labeling for dynamic objects (see Sec \ref{sec:model:generation_of_LiDAR_based_label}), our approach outperforms the baseline and sometimes even the LiDAR DGM.
		
		
	\section{CONCLUSIONS}
	\addtolength{\textheight}{-0.3cm}   
	



	This paper proposed a single-frame deep ISM to generate measurement grids in polar coordinates. The learned RADAR measurement grids combined with Doppler velocity measurements are further used to generate dynamic grid maps (DGMs). Experiments have verified that our deep ISM is independent of sensor mounting. This enables flexibly using the method for additional sensors of the same type without retraining the neural network. We also verified that no pre-aggregation of RADAR frames is the best choice in our architecture. Moreover, experimental results both in the level of measurement grid and DGM indicate that our approach outperforms the geometric ISM in representing static obstacles (guardrails), dynamic objects (vehicles) and free spaces in real-world highway driving scenarios. Regarding dynamic estimates, our RADAR-only approach even occasionally outperforms DGMs generated from high-fidelity LiDARs.
 \section{ACKNOWLEDGEMENT}
The authors would like to especially thank Linda Schubert for her helpful support with the hardware setup and Stefan Müller for the sensor data collection.
	\bibliographystyle{IEEEformats/IEEEtran} 
	\bibliography{IEEEformats/IEEEabrv,IEEEmyref}
\end{document}